\documentclass{article}

\usepackage[preprint]{neurips_2025}

\usepackage[utf8]{inputenc} 
\usepackage[T1]{fontenc}    
\usepackage{hyperref}       
\usepackage{url}            
\usepackage{booktabs}       
\usepackage{amsfonts}       
\usepackage{nicefrac}       
\usepackage{microtype}      
\usepackage{xcolor}         

\usepackage{colortbl}
\definecolor{deepgreen}{rgb}{0.0, 0.5, 0.0}

\usepackage{pifont} 

\usepackage[most]{tcolorbox}
\usepackage{caption}
\usepackage{xcolor}
\usepackage{graphicx}
\usepackage{amsmath}
\usepackage[utf8]{inputenc}

\newtcolorbox[list inside=prompt,auto counter,number within=section]{prompt}[1][]{
    colbacktitle=black!60,
    coltitle=white,
    fontupper=\footnotesize,
    boxsep=5pt,
    enhanced,
    left=0pt,
    right=0pt,
    top=0pt,
    bottom=0pt,
    boxrule=1pt,
    breakable,
    #1
}

\usepackage{makecell}

\newcommand{\modelName}{DentalGPT}
\newcommand{\iconPubmed}{\includegraphics[height=0.65em]{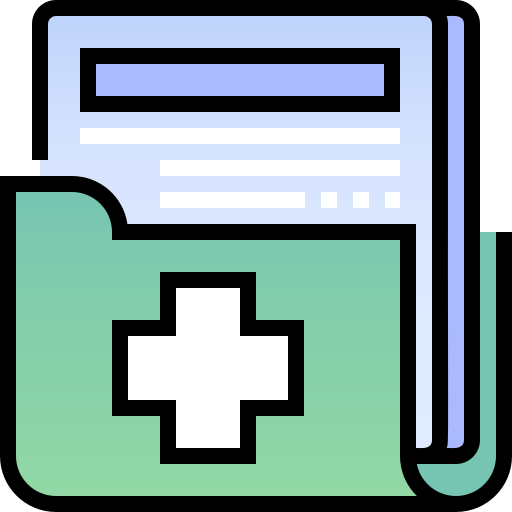}}
\newcommand{\iconImagelabel}{\includegraphics[height=0.65em]{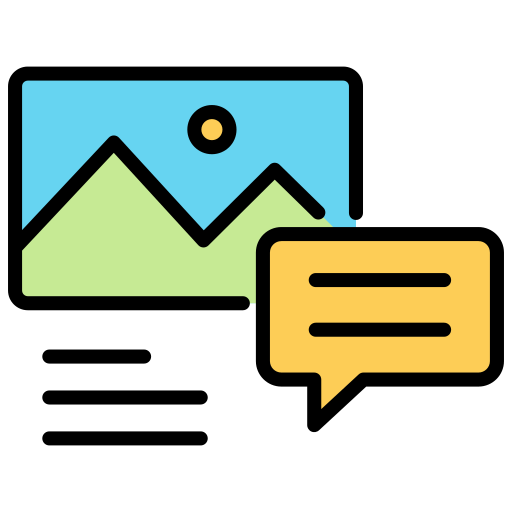}}
\newcommand{\iconBounding}{\includegraphics[height=0.65em]{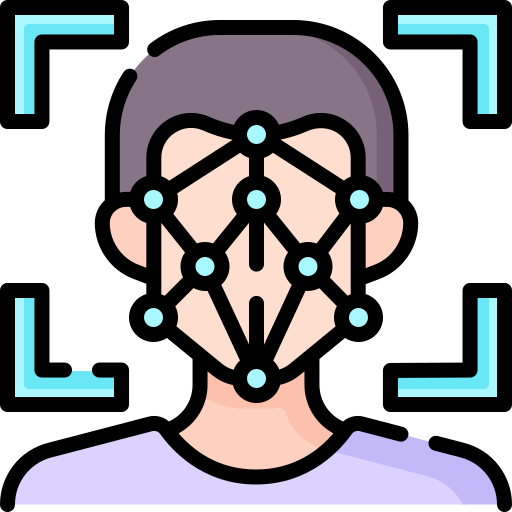}}
\newcommand{\iconDentist}{\includegraphics[height=0.65em]{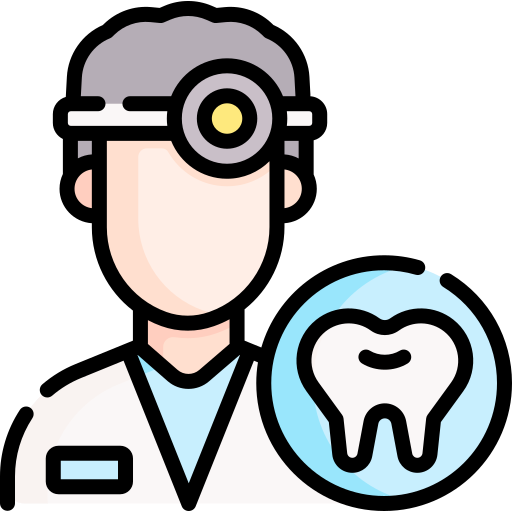}}

\newtcolorbox{AIbox}[2][]{aibox,title=#2,#1}
\definecolor{colorA}{rgb}{0.831, 0.831, 0.902} 
\definecolor{colorB}{rgb}{0.839, 0.953, 0.988} 
\definecolor{colorC}{rgb}{0.847, 0.902, 0.878} 
\definecolor{colorD}{rgb}{0.945, 0.816, 0.780} 
\definecolor{colorE}{rgb}{1.000, 0.965, 0.867} 
\definecolor{customblue}{rgb}{0.1216, 0.4667, 0.7059}
\definecolor{customgreen}{rgb}{0.1725, 0.6275, 0.1725}
\definecolor{customred}{rgb}{0.8392, 0.1529, 0.1569}

\title{\modelName: Incentivizing Multimodal Complex Reasoning in Dentistry}

\author{
\textbf{Zhenyang Cai$^{2\dagger}$, Jiaming Zhang$^{1\dagger}$, Junjie Zhao$^{5,6\dagger}$, Ziyi Zeng$^2$, Yanchao Li$^2$, Jingyi Liang$^2$}\\
\textbf{Junying Chen$^2$, Yunjin Yang$^2$, Jiajun You$^{2,4}$, Shuzhi Deng$^{1}$, Tongfei Wang$^{1}$} \\
\textbf{Wanting Chen$^{1}$, Chunxiu Hao$^{1}$, Ruiqi Xie$^{1}$, Zhenwei Wen$^{5}$, Xiangyi Feng$^4$}\\
\textbf{Zou Ting$^{1}$, Jin Zou Lin$^{1}$, Jianquan Li$^4$, Guangjun Yu$^{2,7}$} \\
\textbf{Liangyi Chen$^{3*}$, Junwen Wang$^{5*}$, Shan Jiang$^{1*}$, Benyou Wang$^{2,7,8,9}$\footnote{The first three authors contributed to this work equally. Liangyi, Junwen, Shan and Benyou are the corresponding authors.}}\\
\textsuperscript{\rm 1} Shenzhen Stomatology Hospital (Pingshan) of Southern Medical University \\
\textsuperscript{\rm 2} The Chinese University of Hong Kong, Shenzhen \textsuperscript{\rm 3} State Key Laboratory of Membrane Biology,\\ Beijing Key Laboratory of Cardiometabolic Molecular Medicine, Institute of Molecular Medicine, \\National Biomedical Imaging Center, School of Future Technology, Peking University\\
\textsuperscript{\rm 4} Freedom AI 
\textsuperscript{\rm 5} Division of Applied Oral Sciences \& Community Dental Care \\ Faculty of Dentistry, The University of Hong Kong \\
\textsuperscript{\rm 6} Beijing Institute of Collaborative Innovation
\textsuperscript{\rm 7} National Health Data Institute, Shenzhen \\ 
\textsuperscript{\rm 8} Shenzhen Loop Area Institute \textsuperscript{\rm 9} Shenzhen Institute of Big Data
}

\begin{document}

\maketitle

\begin{abstract}
Reliable interpretation of multimodal data in dentistry is essential for automated oral healthcare, yet current multimodal large language models (MLLMs) struggle to capture fine-grained dental visual details and lack sufficient reasoning ability for precise diagnosis. To address these limitations, we present \textbf{DentalGPT}, a specialized dental MLLM developed through high-quality domain knowledge injection and reinforcement learning. Specifically, the largest annotated multimodal dataset for dentistry to date was constructed by aggregating over 120k dental images paired with detailed descriptions that highlight diagnostically relevant visual features, making it the multimodal dataset with the most extensive collection of dental images to date. Training on this dataset significantly enhances the MLLM’s visual understanding of dental conditions, while the subsequent reinforcement learning stage further strengthens its capability for multimodal complex reasoning. Comprehensive evaluations on intraoral and panoramic benchmarks, along with dental subsets of medical VQA benchmarks, show that DentalGPT achieves superior performance in disease classification and dental VQA tasks, outperforming many state-of-the-art MLLMs despite having only 7B parameters. These results demonstrate that high-quality dental data combined with staged adaptation provides an effective pathway for building capable and domain-specialized dental MLLMs.
\end{abstract}


\section{Introduction}
\label{sec:intro}

Dental healthcare is an essential area of public health, yet the workload of dental professionals continues to increase each year~\cite{cdc2024oralhealth, jiang2024dentists, zheng2025comparative, lilford2025supply}. To support both clinicians and patients, multimodal large language models (MLLMs)~\cite{liu2023visual, qwen2.5vl, team2023gemini, chen2024internvl} capable of interactive communication through dialogue have recently attracted significant attention, offering new possibilities for intelligent dental care. However, despite their promising performance in general medical applications~\cite{awadalla2023openflamingo, li2023llavamed, wu2023generalist, wu2024pmc, huatuogptv, su2025gmai, xu2025lingshu}, current MLLMs still face notable limitations when dealing with more specialized medical imaging problems, such as images in dentistry.

After extensive observation, we found that although current MLLMs can sometimes identify relevant features in dental images, they often fail to extract the necessary information and reason over it to produce correct answers. This exposes two challenges when applying MLLMs to dentistry. First, current MLLMs lack sufficient visual understanding of dental images. This limitation prevents them from effectively using their knowledge to perform reliable diagnostic reasoning. Second, although recent MLLMs exhibit strong complex reasoning capabilities that have led to substantial gains in complex image understanding tasks~\cite{su2025gmai,qwen3vl,glm4v,zhang2025med,pan2025medvlm}, our quantitative results show that such reasoning provides only marginal improvements in dentistry. This suggests that there remains significant room for advancing complex reasoning specifically tailored to dental diagnosis.

To address these challenges, we injected dental knowledge into the MLLM and strengthened its ability for complex reasoning. Specifically, we collected a large number of dental images with textual descriptions or labels from online sources and combined them with professionally annotated images from dental hospitals. Together, these resources form a dataset, which includes over 120k dental images with detailed descriptions and additional QA pairs for specific downstream tasks. The descriptions highlight diagnostically relevant cues in the image, helping the model connect visual features with its textual understanding of the conditions, while the QA data further enhances its downstream task performance. After the enhancement of multimodal understanding, a reinforcement learning (RL) stage with the Group Relative Policy Optimization (GRPO)~\cite{grpo} algorithm was applied, guiding the model to use the added knowledge to explore more explanatory solutions for dental questions. This process resulted in \textbf{DentalGPT}, the first specialized dental MLLM equipped with complex reasoning capabilities.

Then, a comprehensive evaluation was conducted to assess DentalGPT's ability in dental image analysis. First, MMOral Bench and a dentistry-focused subset of medical VQA benchmarks were curated to evaluate the model’s generalization ability. Second, professionally annotated intraoral and panoramic images were used to assess its competence in identifying specific dental diseases. The results show that DentalGPT, despite having only 7 billion parameters, surpasses existing MLLMs in dental image understanding and question answering, highlighting its efficiency and domain specialization.

Beyond overall performance, the impact of the two training stages was examined through in-depth analysis, revealing several key findings. The training data exhibited higher knowledge density and greater professional quality compared with \texttt{GPT-5}–distilled data, indicating their advantage for domain learning. Based on this stronger data foundation, the Multimodal Understanding Enhancement stage was shown to substantially enrich the model's dental knowledge and improve its performance across diverse image analysis tasks. These gains were further amplified during reinforcement learning, resulting in higher disease classification accuracy and more professional identification of diagnostically relevant visual cues. Together, these results demonstrate that high-quality dental data, combined with staged training, plays a critical role in shaping DentalGPT's specialized capabilities.

In conclusion, our contributions are: 1) Introduce \textbf{DentalGPT}, a specialized dental MLLM equipped with advanced dental image analysis and complex reasoning capabilities to interpret fine visual details and associate them with related conditions. 2) Curate a large-scale VQA dataset containing the largest collection of dental images to date with detailed descriptions of diagnostically relevant visual features. 3) Demonstrate that DentalGPT achieves strong performance on multiple disease classification and VQA tasks in dentistry, surpassing many larger MLLMs.


\section{On Incentivizing Multimodal Complex Reasoning in Dentistry}

\subsection{Motivations for Multimodal Complex Reasoning in Dentistry}
\label{sec:motivation}

Dentistry is a key medical field that relies on clinicians analyzing patients’ imaging data and communicating with them for diagnosis, yet even leading MLLMs with strong general multimodal abilities still fall far behind professional dentists in multimodal diagnosis.

\begin{figure}[ht!]
    \centering
    \includegraphics[width=\linewidth]{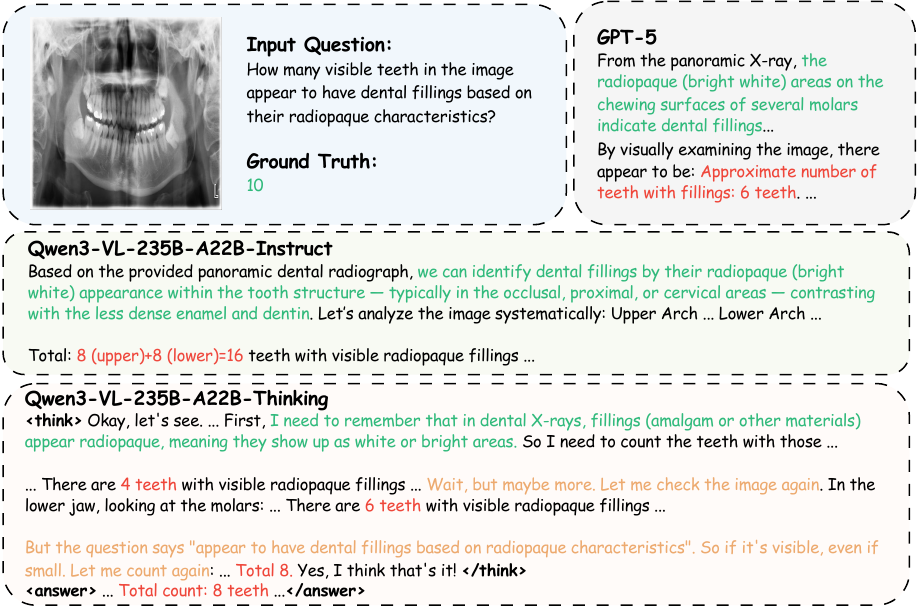}
    \caption{Examples of top-tier general-purpose MLLMs analyzing a dental image task. Red indicates incorrect analysis, green indicates correct analysis, and orange highlights reflective turns in the complex reasoning process.}
    \label{fig:sec2-examples}
\end{figure}

\paragraph{Case Study} To investigate this phenomenon, we conducted a case study on a dental multimodal task using one leading commercial MLLM as well as both the \textit{Instruct} and \textit{Thinking} mode of a state-of-the-art open-source MLLM (Figure~\ref{fig:sec2-examples}). After analyzing the model outputs, we found that although they could identify the relevant visual features to be counted, they still produced incorrect answers. Notably, by examining the reasoning trajectory of \texttt{Qwen3-VL-235B-A22B-Thinking}, we observed that it repeatedly reflected on and revised its own intermediate conclusions. Through this type of complex reasoning pattern, the model gradually approached the correct answer step by step. Although it still failed to produce the correct final prediction, this behavior highlights the potential of applying complex reasoning strategies to achieve more accurate multimodal diagnosis in dentistry.

\begin{figure}[ht!]
    \centering
    \includegraphics[width=0.8\linewidth]{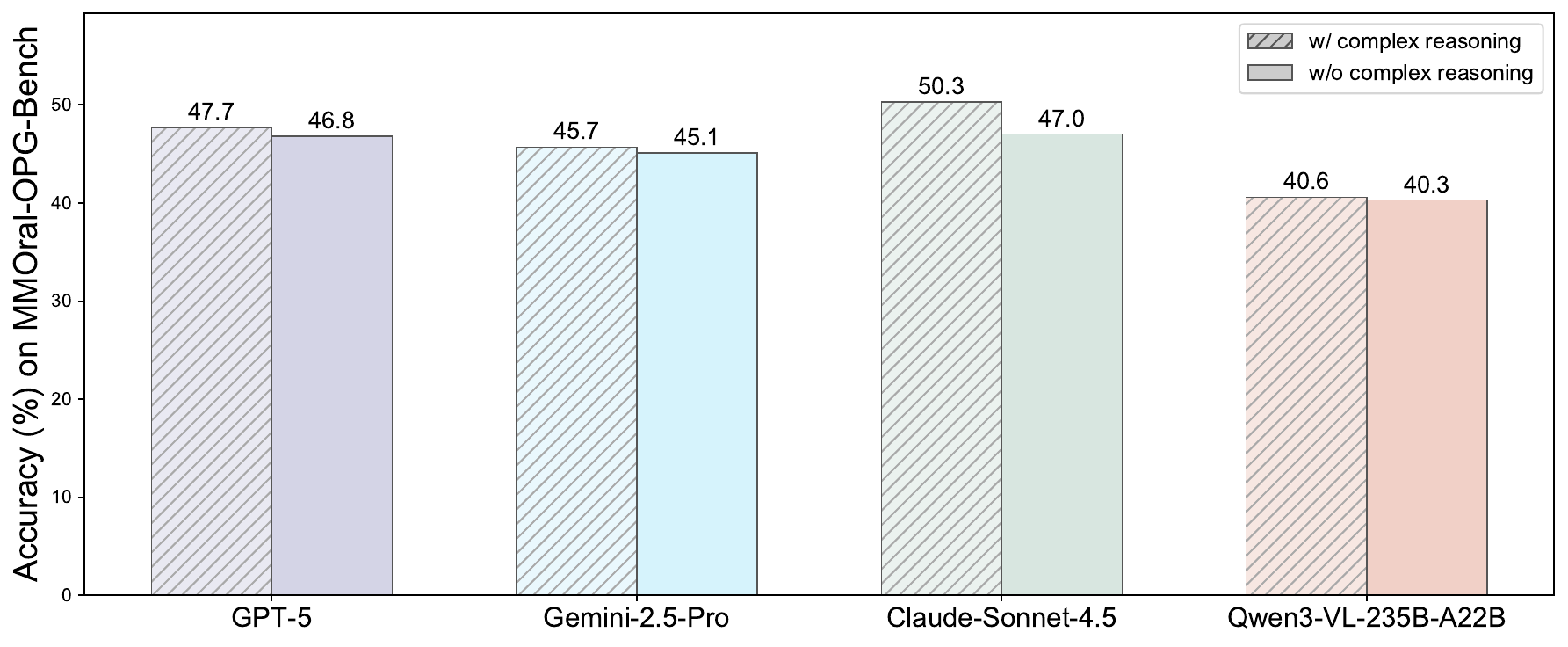}
    \caption{Accuracy (\%) of MLLMs \textit{with and without the complex reasoning mode} on the MMOral-OPG-Bench.}
    \label{fig:sec2-quanti}
\end{figure}

\paragraph{Quantitative Experiment} Furthermore, to quantitatively assess model performance on dental tasks and examine whether complex reasoning provides additional benefits, We evaluated leading MLLMs~\footnotemark{} \textit{with and without the complex reasoning mode} using the existing MMOral-OPG-Bench~\cite{hao2025oralgpt}. The results in Figure~\ref{fig:sec2-quanti} show that both models achieve higher scores in their \textit{complex reasoning} mode, further demonstrating the potential of complex reasoning in dentistry.
\footnotetext{
\textbf{Complex reasoning mode:} GPT5-2025-08-07, Gemini-2.5-Pro-Thinking,
Claude-Sonnet-4-5-20250929-Thinking, Qwen3-VL-235B-A22B-Thinking.
\textbf{Without complex reasoning mode:} GPT5-chat-2025-08-07,
Gemini-2.5-Pro-NoThinking, Claude-Sonnet-4-5-20250929,
Qwen3-VL-235B-A22B-Instruct.
}

\subsection{Challenges for Multimodal Complex Reasoning in Dentistry}

Prior works~\cite{su2025gmai,qwen3vl,glm4v,zhang2025med,pan2025medvlm} have shown that complex reasoning can significantly improve MLLM performance in mathematics, medicine and other complex image understanding tasks. However, the results in Section~\ref{sec:motivation} indicate that the benefits of complex reasoning for dentistry-related multimodal tasks remain limited. Thus, after a closer analysis of the quantitative results, we conclude the following challenges:

\paragraph{Challenge I: Limited Multimodal Understanding in Dentistry} MLLMs in Section~\ref{sec:motivation} perform poorly on MMOral-OPG-Bench (None of them exceed 60\% accuracy). The foundation of complex visual reasoning is a model's ability to accurately interpret dental images, and a straightforward way to enhance this ability is to inject visual domain knowledge through large-scale data~\cite{liu2023visual}. However, existing dental vision–language datasets remain very limited; for instance, only about 0.3 percent of PubMedVision~\cite{huatuogptv} images involve teeth. To address this gap and support greater multimodal understanding in dentistry, we introduce a multimodal dataset with 120k dental images.

\paragraph{Challenge II: Limited Complex Reasoning in Dentistry} MLLMs employing complex reasoning refine and verify their intermediate conclusions, allowing them to gradually approach the correct answer. As shown in Section~\ref{sec:motivation}, current MLLMs achieve only limited gains on dentistry-related tasks with the complex reasoning. However, prior studies~\cite{su2025gmai,zhang2025med,pan2025medvlm} about medical MLLMs have shown that task-specific reinforcement learning can strengthen these reasoning abilities and improve downstream performance. This motivates us to explore domain-specific reinforcement learning to further enhance complex reasoning and improve performance in dentistry.


\section{\modelName: From Multimodal Understanding to Complex Reasoning }
\label{sec:dentalgpt}

\begin{figure}[ht!]
    \centering
    \includegraphics[width=1\linewidth]{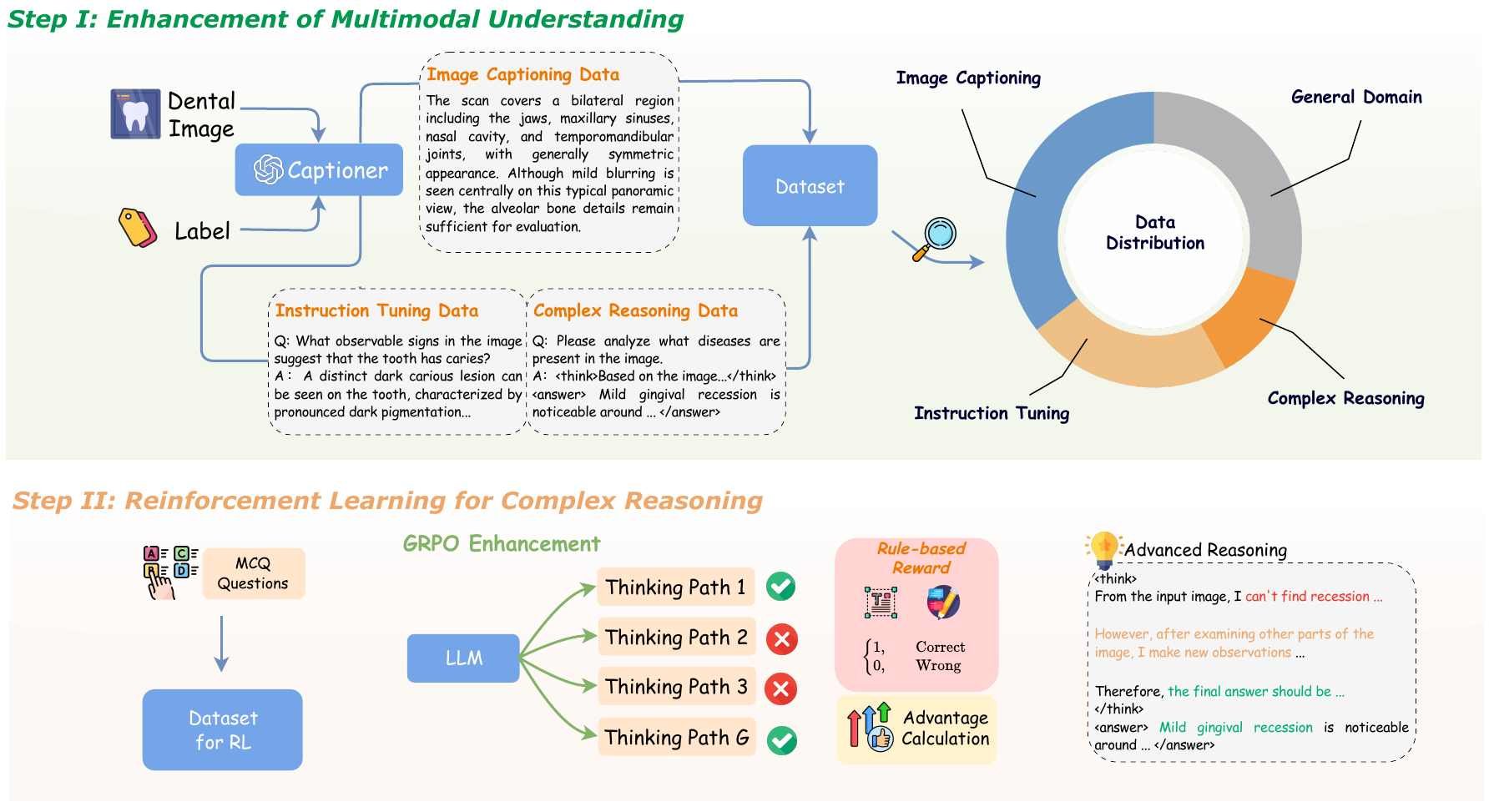}
    \caption{The 2-stage process of building DentalGPT. Multimodal Understanding Enhancement stage uses a large dataset to align the model's medical knowledge with its multimodal understanding and prepare it for downstream tasks; Reinforcement Learning then strengthens complex reasoning ability.}
    \label{fig:dentalgpt-main}
\end{figure}

To overcome the limited multimodal understanding and basic reasoning patterns of existing MLLMs in dental imaging, DentalGPT is proposed as a dentistry-specific MLLM built through a 2-stage training process (Figure~\ref{fig:dentalgpt-main}). Stage I focuses on multimodal understanding enhancement to strengthen the MLLM’s understanding of dental images, while Stage II leverages this enhanced ability through reinforcement learning to improve complex reasoning.

\subsection{Stage I: Enhancement of Multimodal Understanding}

To address the limited multimodal understanding of existing MLLMs in dentistry, this stage enriches the model with high-quality dental knowledge using large-scale, professionally curated image–text data. \texttt{GPT-5} was used to generate detailed descriptions, forming a specialized multimodal dataset that strengthens fine-grained visual understanding and basic downstream task performance.

\paragraph{Data Composition} At this stage, the collected images and their corresponding labels are organized into several types of training data. First, \textbf{Image Captioning} data is used to train the model to clearly and comprehensively describe dental images, helping align textual representations with real-world dental concepts. Second, \textbf{Instruction Tuning} data consists of a large number of question–answer samples, enabling the model to understand user intent and respond appropriately. Third, \textbf{Complex Reasoning} data includes multi-step and reflective thinking examples, serving as a foundation for subsequent reinforcement learning to enhance multimodal complex reasoning capabilities. Finally, general-domain data~\cite{liu2023visual,LLaVA-OneVision-1.5,chen2024sharegpt4v} is incorporated to maintain the model’s ability to understand both images and text beyond dental scenarios, preventing overfitting to dental-specific tasks.

\paragraph{Training Settings} The model was then trained on this dataset for two epochs with a batch size of 256 and a learning rate of $2 \times 10^{-5}$. All parameters were fully updated during training, with the first 5\% of steps allocated for learning-rate warmup.

\subsection{Stage II: Reinforcement Learning for Complex Reasoning}

After gaining new knowledge, the MLLM must learn to apply it for improved complex reasoning in multimodal diagnosis. Recent works such as \texttt{DeepSeek-R1}~\cite{deepseekr1} and \texttt{GPT-o1}~\cite{gpto1} show that reinforcement learning can encourage long chain-of-thought generation and enhance reasoning quality. Following this paradigm, we adopt GRPO to optimize the reasoning process of \textit{DentalGPT}.

\paragraph{Data Composition} To achieve this goal, we selected a set of dental images that were not used during the Stage I to construct a new dataset. Based on the original labels and their label sets, we generated multiple-choice questions with correct answers, enabling rule-based correctness checking, which is crucial for reward computation in GRPO.

\paragraph{Training Strategy}
During this stage, we employ the GRPO algorithm to enhance the model's reasoning ability on dental multiple-choice tasks. GRPO evaluates relative advantages within sampled groups of responses, enabling efficient optimization without requiring a value network.

\begin{itemize}
    \item \textbf{Sampling Action Groups.}  
    For each input state $s = (I, Q)$, where $I$ is the dental image and $Q$ the question, a fixed prompt is appended to require the model to generate reasoning inside \texttt{\textless think\textgreater} tags and produce the final answer within \texttt{\textless answer\textgreater} tags. GRPO then samples a group of candidate responses:
    \begin{equation*}
        a_i \sim \pi_{\theta}(a \mid I, Q),\quad i = 1,\dots,G.
    \end{equation*}
    This promotes diverse reasoning paths and prevents premature convergence.

    \item \textbf{Reward Design.}  
    Each sampled answer $a_i$ is evaluated with a composite reward:
    \begin{equation*}
        R(a_i) = 0.1\,R_{\text{format}}(a_i) + 0.9\,R_{\text{acc}}(a_i).
    \end{equation*}
    \begin{itemize}
        \item \textit{Format Reward:}  
        Ensures adherence to the required template:
        \begin{equation*}
        R_{\text{format}}(a_i) =
        \begin{cases}
            1, & \text{if formatted correctly},\\
            0, & \text{otherwise}.
        \end{cases}
        \end{equation*}

        \item \textit{Accuracy Reward:}  
        For multiple-choice questions:
        \begin{equation*}
        R_{\text{acc}}(a_i) =
        \begin{cases}
            1, & \text{if the prediction is correct},\\
            0, & \text{otherwise}.
        \end{cases}
        \end{equation*}
    \end{itemize}

    \item \textbf{Policy Update.}  
    Rewards within the group are normalized to produce relative advantages:
    \begin{equation*}
        A_i = 
        \frac{R(a_i) - \mathrm{mean}(R(a_1),\dots,R(a_G))}
        {\mathrm{std}(R(a_1),\dots,R(a_G))}.
    \end{equation*}
    GRPO then updates the policy to reinforce high-advantage actions while constraining deviations from the reference policy via KL regularization.
\end{itemize}

\paragraph{Training Settings} During GRPO optimization, the model was optimized using grouped rollouts with a sampling size of 10 responses per prompt. Training was conducted with a rollout batch size of 256 and a learning rate of $1\times10^{-6}$. The optimization ran for 5 epochs, and the maximum response length was capped at 8192 tokens to accommodate long CoT reasoning. This configuration ensured stable exploration within each action group while maintaining sufficient capacity for detailed reasoning outputs.


\section{Data Engineering}

This section elaborates on the data foundation supporting \textbf{DentalGPT}, including the sources of the training data and the quality control processes used to ensure high-quality dental images; additionally, we analyze the overall data quality to ensure that its completeness, knowledge depth, and safety remain at a leading level.

\subsection{Data Collection}
\begin{figure}[ht!]
    \centering
    \includegraphics[width=0.9\linewidth]{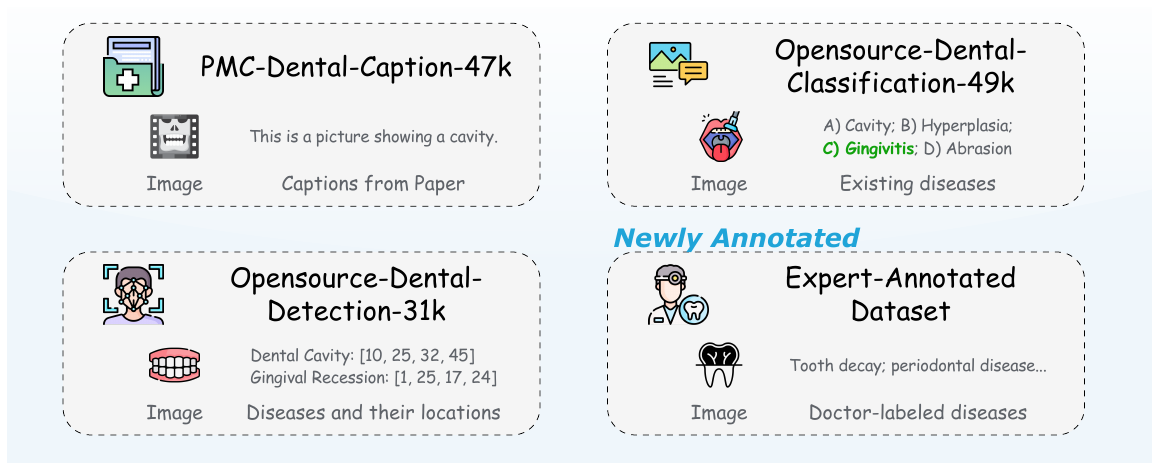}
    \caption{Annotation examples from different dental image collections.}
    \label{fig:datasetcollection}
\end{figure}

Before improving an MLLM’s understanding and reasoning on dental images, it is necessary to collect a sufficiently large set of training samples (examples are shown in Figure~\ref{fig:datasetcollection}), which provides the foundation for the model to effectively use its knowledge during image interpretation~\cite{liu2023visual}.

\subsubsection{Existing Annotated Data}

To efficiently obtain a sufficiently large number of dentistry-specific multimodal datasets, we first sourced datasets from a variety of open-source platforms. By leveraging certified, high-quality dental image datasets and existing literature, we aim to enrich the model's understanding of relevant radiographic data.

\paragraph{\iconPubmed~PMC-Dental-Caption-47k}
PubMed Central (PMC) is a publicly accessible biomedical repository that hosts a vast collection of peer-reviewed medical publications. It is considered a reliable and widely used data source in previous research. From PMC, we filtered a large number of dental images and retained the associated captions and labels provided within the original papers. This rich textual context is expected to provide valuable information for enhancing visual understanding and facilitating the integration of dental domain knowledge into our model.

\paragraph{\iconImagelabel~Opensource-Dental-Classification-49k}  
To further leverage image datasets that have previously been used to train classification models, we collected a wide range of dental-related classification datasets and consolidated them into a larger corpus of dental images with corresponding labels. Specifically, each image is associated with one or more disease labels; for multi-class or multi-label datasets, we also retain the negative labels so that all available, clinically validated annotations can be fully exploited. This unified resource supports MLLMs in better aligning common dental disease categories with key feature identification.

\paragraph{\iconBounding~Opensource-Dental-Detection-31k}
We also collected a number of datasets previously utilized for dental lesion localization tasks, in which each image is annotated with one or more lesion instances together with their spatial coordinates. Although our model is not explicitly trained to predict bounding boxes, such annotations provide MLLMs with implicit spatial cues and lesion counts, thereby supporting the model’s capability to understand spatial relationships and quantify dental abnormalities.

\subsubsection{Newly Annotated Data}

Building upon the open-source datasets described above, we observed that although they contain a considerable number of annotations, the diagnostic focus is predominantly centered on a few common dental conditions. From a clinical perspective, however, there exist additional critical abnormalities and visual manifestations that warrant greater attention yet are underrepresented in existing data sources. To address this gap, we expanded the diagnostic label set and further curated a new subset of dental images, which were annotated by certified dental experts with an emphasis on clinically significant conditions and indicative visual features.

\paragraph{\iconDentist~Expert-Annotated Dataset}
We collected some dental images from internet sources, hospital imaging archives, and publicly available datasets. After removing duplicates and low-resolution images, we curated a candidate dataset that was subsequently annotated by professional dental clinicians. To ensure annotation quality, a strict cross-validation mechanism was applied with different levels of control over data variation. For the training set, annotations from dentists with a cross-validation agreement rate below 85\% were filtered out, while the remaining labels were retained for constructing caption-style training data. For the test set, a more rigorous protocol was adopted: each sample was annotated by at least two dentists, and only those with consistent diagnostic results were preserved. This dataset equips the model with specialized clinical knowledge and a broader ability to recognize dental diseases and clinically relevant signs.

\subsection{Data Curation}

To enrich the model’s ability in image understanding, instruction following, and task-specific dental reasoning, we further curated and refined all collected data using \texttt{GPT-5}. This curation step enables the model to learn not only visual concepts but also how to interpret clinical instructions and produce structured diagnostic outputs. The overall process is outlined as follows:

\paragraph{Image Captioning} This component focuses on enhancing the model's ability to capture diagnostically relevant visual details in dental images. \texttt{GPT-5} was instructed to describe all observable features that may aid diagnosis while referencing the original descriptions and labels, and to avoid any diagnostic assumptions. These observation-based captions were then paired with predefined questions and answers to form a caption-based VQA subset, improving the model's interpretation of dental images and reducing visual information gaps.

\paragraph{Instruction Tuning} To enhance performance on different downstream tasks, \texttt{GPT-5} was prompted to generate questions based on the collected images and their descriptions, simulating real diagnostic scenarios. Additionally, several vision models were used to annotate a portion of the datasets, and only high-confidence annotations were retained. \texttt{GPT-5} then refined these annotations and transformed them into structured question–answer pairs, further improving the model's ability to handle dental question answering.

\paragraph{Complex Reasoning} To support the subsequent reinforcement learning stage, complex chain-of-thought data were constructed by \texttt{GPT-5} following the \texttt{HuatuoGPT-o1}~\cite{chen2024huatuogpto1} methodology. Fixed reasoning templates were added to activate the model's basic multi-step reasoning ability, enabling it to analyze before answering.

After these steps were completed, \texttt{GPT-5-mini} was used to perform a secondary verification of all data. Entries that diverged from the original image descriptions or labels were removed, further ensuring data accuracy and producing the training dataset.

\subsection{Quality Assessment}

As described in the methodology, the training dataset of DentalGPT was generated by \texttt{GPT-5} while referencing existing image labels or descriptions to minimize hallucinations and ensure professional domain knowledge injection. To validate the effectiveness of this approach, five evaluation dimensions were defined, and a randomly sampled set of 3,000 entries was assessed and compared against data obtained through direct \texttt{GPT-5} distillation. To ensure fairness, all comparisons were evaluated using \texttt{Gemini-2.5-Pro} as the judge.

\paragraph{Evaluation Setup} Specifically, the dataset was evaluated across the following five dimensions: 1) \colorbox{colorA!60}{\textit{Description Completeness}}: Whether all observable visual details in the image are thoroughly described, with particular attention to features that may contribute to dental diagnosis. 2) \colorbox{colorB!60}{\textit{Terminology Consistency}}: Whether professional dental terminology is used correctly and consistently throughout the description. 3) \colorbox{colorC!60}{\textit{Content Safety}}: Whether the content adheres to medical ethics and safety standards, avoiding sensitive, discriminatory, misleading, or inappropriate statements. 4) \colorbox{colorD!60}{\textit{Text–Image Consistency}}: Whether the textual description is well written and accurately aligned with the corresponding image content. 5) \colorbox{colorE!80}{\textit{Knowledge Depth}}: Whether the description demonstrates an appropriate level of dental knowledge.

\texttt{Gemini-2.5-Pro} was asked to score each dimension on a scale of 1 to 5, and the final dataset quality was reported using the average score across all evaluated samples.

\begin{figure}[ht!]
    \centering
    \includegraphics[width=0.6\linewidth]{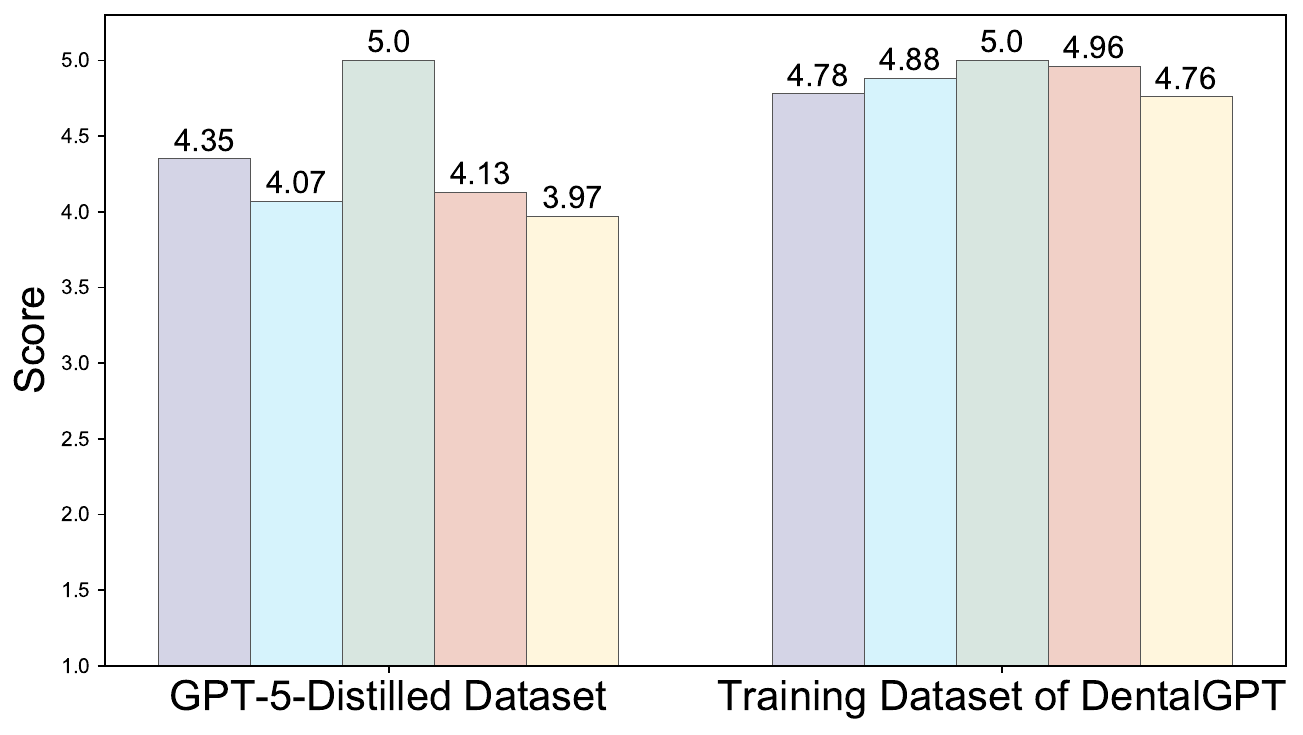}
    \caption{\texttt{Gemini-2.5-Pro}'s multi-dimensional evaluation of \texttt{GPT-5}–distilled data and the training dataset of DentalGPT (scores range from 0 to 5). For each dataset, the color blocks from left to right represent \colorbox{colorA!60}{\textit{Description Completeness}}, \colorbox{colorB!60}{\textit{Terminology Consistency}}, \colorbox{colorC!60}{\textit{Content Safety}}, \colorbox{colorD!60}{\textit{Text–Image Consistency}}, and \colorbox{colorE!80}{\textit{Knowledge Depth}}.}
    \label{fig:indepth-dataquality}
\end{figure}

\paragraph{Results Analysis} Results are shown in Figure~\ref{fig:indepth-dataquality}, our training dataset demonstrates clear advantages over the directly distilled version across multiple evaluation metrics. It can be observed that the data generated with label references shows the most significant improvements in \textit{terminology consistency} and \textit{knowledge depth}, and also achieves notable gains in the completeness of visual detail descriptions. Interestingly, \texttt{Gemini-2.5-Pro} assigns a perfect score for safety to both datasets, indicating that \texttt{GPT-5}–generated data perform exceptionally well in medical safety, avoiding harmful diagnostic suggestions and providing timely guidance to reduce potential risks.

In conclusion, the results indicate that our dataset, by leveraging annotations from public datasets and descriptions from academic literature, provides more comprehensive and more professional knowledge injection for the model. Such a foundation ensures substantial improvement in the performance of DentalGPT.

\section{Benchmark Design and Construction}
\label{sec:benchmark}

To comprehensively evaluate the model, existing benchmarks containing dental images were first used to assess its performance. Additionally, a large set of dental images was collected and annotated with disease labels by professional dentists, ensuring clinical validity and allowing further assessment aligned with expert consensus.

\subsection{Existing Benchmarks}
We also conducted comprehensive evaluations on several open-source medical VQA benchmarks to assess the model's performance in more scenarios.

\quad \textbf{(1) \textit{MMOral-OPG-Bench}} MMOral~\cite{hao2025oralgpt} contains multiple panoramic dental images with high-quality expert annotations and spans five clinically grounded dimensions, offering a thorough assessment of an MLLM's panoramic X-ray understanding. We use its open-ended test split to more directly and intuitively evaluate model performance.

\quad \textbf{(2) \textit{DentalBench-Mixed}} was further constructed by strictly filtering tooth-related images from existing medical VQA benchmarks. Specifically, the data were sourced from PMC-VQA~\cite{zhang2023pmcvqa}, OmniMedVQA~\cite{hu2024omnimedvqa}, and MedXpertQA-MM~\cite{zuo2025medxpertqa}, which are widely used for evaluating the capabilities of medical MLLMs. The dental-relevant portions of these benchmarks were extracted to form the \textit{DentalBench-Mixed} dataset, enabling targeted assessment of models on dental image understanding.

\subsection{Expert-annotated Benchmarks}

\paragraph{Annotation Workflow}
To maximize data diversity and ensure clinically reliable model outputs, we collected dental images from multiple sources and invited professional dentists to perform expert annotations. Based on advice from dentists, we defined a set of commonly observed dental diseases and clinically relevant signs that are either diagnostically important themselves or serve as auxiliary evidence in clinical reasoning. These labels were then used to assess whether the model could correctly identify key visual cues in dental images.

\begin{figure}[ht!]
    \centering
    \includegraphics[width=0.4\linewidth]{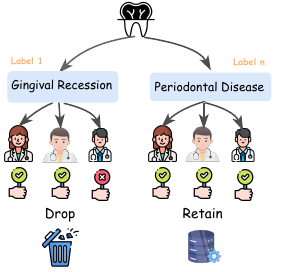}
    \caption{Cross-validation workflow used for benchmark labeling.}
    \label{fig:crossvalidation}
\end{figure}

To ensure annotation reliability, a rigorous cross-validation workflow (shown in Figure~\ref{fig:crossvalidation}) was implemented. Each image was independently annotated by at least two dentists, who were asked to select all clinically relevant labels present in the image. Dentists were also allowed to mark an image as “uncertain” when image quality or lighting conditions made visual assessment unreliable, thereby reducing the risk of forced or ambiguous judgments. After annotation, labels with a low agreement rate (below 85\%) were removed, as they indicate visual patterns that cannot be consistently identified from the images alone. Furthermore, any labels with disagreement between annotators were filtered out, and only those with full consensus were retained to guarantee high diagnostic reliability.

After the expert annotation process, only high-quality and clinically reliable labels were retained to form the final annotated dataset. This curated dataset serves as the foundation for constructing our expert-annotated benchmarks and enables a rigorous evaluation of the model’s multimodal diagnostic capability.

\paragraph{Benchmark Composition}

Based on the annotations workflow described above, we construct three benchmarks to comprehensively evaluate the multimodal diagnosis performance of DentalGPT. These subsets cover both clinical and in-the-wild imaging conditions and span key dental diseases and clinically relevant signs that dentists frequently observe in real practice. Each benchmark targets different imaging modalities or usage scenarios, enabling a thorough assessment of the model's reliability, generalization ability, and robustness across diverse dental settings.

\begin{figure}[ht!]
    \centering
    \includegraphics[width=0.9\linewidth]{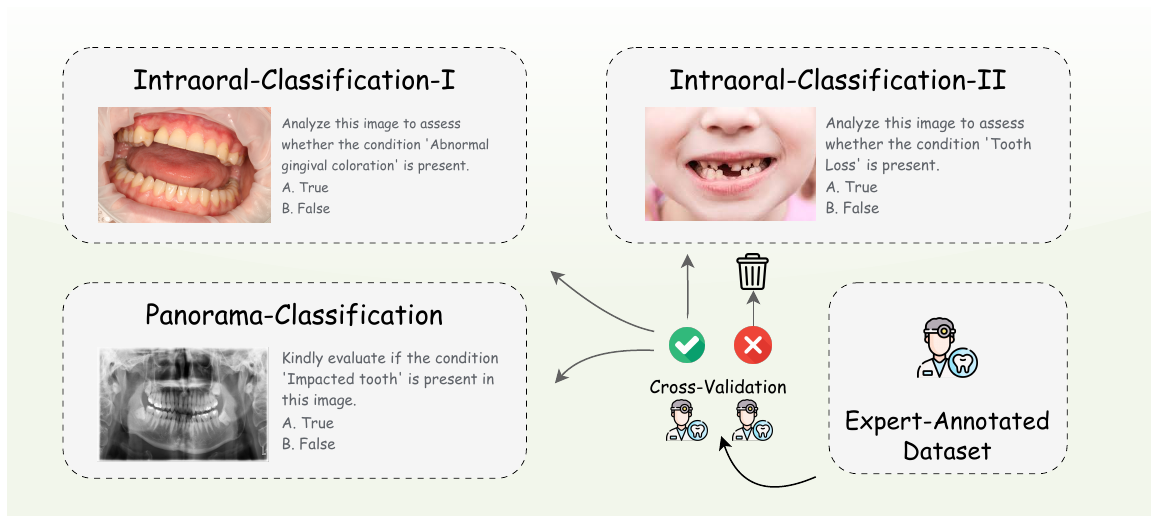}
    \caption{Examples of Expert-annotated Benchmarks}
    \label{fig:benchexample}
\end{figure}

\quad \textbf{(1) \textit{Intraoral-Classification-I}}: This benchmark is sourced from the existing dental image classification dataset AlphaDent~\cite{alphadent}, which contains intraoral photographs collected by multiple clinical dentists under standardized lighting and imaging conditions. The dataset provides high-quality clinical views and includes expert annotations for ten conditions: \textit{tooth discoloration, abnormal gingival coloration, gingival recession, dental caries, tooth pigmentation, tooth defect or loss, tooth loss, dental calculus, abnormal tooth morphology, and abnormal gingival morphology}.

\quad \textbf{(2) \textit{Intraoral-Classification-II}}: This benchmark is composed of intraoral images collected from internet searches. Most images were taken directly by patients, resulting in diverse lighting conditions and shooting angles. After removing duplicates with the training set, professional dentists annotated seven conditions: \textit{tooth pigmentation, abnormal gingival coloration, dental calculus, tooth loss, dental caries, abnormal gingival morphology, and gingival recession. This dataset reflects the model's generalization ability to non-clinical, in-the-wild photographs}.

\quad \textbf{(3) \textit{Panorama-Classification}}: This benchmark consists of real clinical panoramic radiographs (X-ray images) collected from hospitals. Unlike intraoral photos captured by conventional cameras, panoramic images reveal structural and pathological features invisible in standard photographs. Professional dentists annotated six categories: \textit{periodontal disease, root canal treatment, tooth defect or loss, jawbone lesion, periapical lesion, and impacted tooth}.

Each benchmark supports multi-label classification across both clinical and in-the-wild scenarios, enabling comprehensive evaluation of multimodal diagnostic ability. To avoid biased evaluation, we further applied strict data balancing strategies: the label distributions and the ratios of positive and negative samples for each category were aligned across subsets. This ensures that accuracy reliably reflect model performance rather than being influenced by label frequency or overrepresented diseases.


\section{Experiment}
\begin{table}[ht!]
\centering
\small
\resizebox{\textwidth}{!}{
\begin{tabular}{l|ccccc|l}
\toprule
\textbf{Model} & \makecell{MMOral\\OPG-Bench} & \makecell{DentalBench\\Mixed} & \makecell{Intraoral\\Classification-I} & \makecell{Intraoral\\Classification-II} & \makecell{Panorama\\Classification} & Avg. \\
\midrule
\rowcolor{colorB!40} \multicolumn{7}{l}{Open-source MLLMs} \\
\midrule
Deepseek-VL2~\cite{wu2024deepseek} & 39.1 & 22.6 & 51.1 & 59.4 & 55.1 & 45.5 \\
Mistral-Large-2512~\cite{mistral2024large} & 41.9 & 48.2 & 50.7 & 58.0 & 44.2 & 48.6 \\
Phi-4-Multimodal-Instruct~\cite{abouelenin2025phi} & 38.5 & 44.4 & 52.2 & 63.3 & 61.5 & 52.0 \\
Ernie-4.5-VL-424B-A47B$^*$~\cite{ernie2025technicalreport} & 45.0 & 51.4 & 58.1 & 65.1 & 44.9 & 52.9 \\
Qwen3-VL-235B-A22B-Instruct~\cite{qwen3vl} & 40.3 & 51.6 & 50.7 & 58.0 & 55.8 & 51.3 \\
Gemma-3-27B-it~\cite{gemma_2025_report} & 42.2 & 43.0 & 51.5 & 61.4 & 59.6 & 51.5 \\
GLM-4.5v$^*$~\cite{glm4v} & 45.7 & 51.4 & 54.8 & 64.7 & 54.5 & 54.2 \\
Qwen3-VL-235B-A22B-Thinking$^*$~\cite{qwen3vl} & 40.6 & 51.6 & 56.7 & 65.7 & 60.3 & 55.0 \\
LLaMA-4-Maverick~\cite{llama4_maverick_2025} & \underline{51.4} & 53.9 & \underline{61.1} & 67.1 & 59.0 & 58.5 \\
\midrule
\rowcolor{colorB!40} \multicolumn{7}{l}{Proprietary MLLMs} \\
\midrule
Claude-Sonnet-4.5~\cite{claude45} & 47.0 & 50.4 & 51.9 & 59.4 & 50.0 & 51.7 \\
Claude-Sonnet-4.5-Thinking$^*$~\cite{claude45} & 50.3 & 53.9 & 55.2 & 66.7 & 55.8 & 56.4 \\
Grok-4.1-Fast & 47.1 & 52.2 & 57.0 & 65.2 & 62.2 & 56.7 \\
Gemini-2.5-Pro-Thinking$^*$~\cite{gemini25} & 45.7 & \textbf{57.4} & 57.0 & 65.2 & \underline{64.1} & 57.9 \\
GPT-4.1~\cite{gpt41} & 47.2 & 51.7 & 60.4 & 70.5 & 61.5 & 58.3 \\
GPT-5$^*$~\cite{gpt5} & 47.7 & 54.3 & 59.3 & \underline{71.0} & 63.5 & \underline{59.2} \\
\midrule
\rowcolor{colorB!40} \multicolumn{7}{l}{DentalGPT and Its Backbone} \\
\midrule
Qwen2.5-VL-7B-Instruct~\cite{qwen2.5vl} & 27.0 & 46.1 & 48.8 & 61.8 & 50.0 & 46.7 \\
\textbf{DentalGPT}$^*$ & \textbf{60.0} & \underline{54.4} & \textbf{64.1} & \textbf{72.9} & \textbf{84.0} & \textbf{67.1} \\
\bottomrule
\end{tabular}
}
\caption{Accuracy (\%) of MLLMs on Dental-related VQA Benchmarks. \textbf{Bold} indicates the best score; \underline{underlines} marks the second-best. $^*$ indicates that the model has activated complex reasoning.}
\label{tab:main-result}
\end{table}

\subsection{Experimental Setup}
\paragraph{Training Settings} DentalGPT is developed on top of the \texttt{Qwen2.5-VL-7B-Instruct}. During both stages of our training pipeline, all parameters of the model are fully updated to ensure comprehensive adaptation to domain knowledge of dentistry and complex reasoning. The complete hyperparameter settings and implementation details for each stage are provided in the respective subsections of Section~\ref{sec:dentalgpt}. All training experiments are performed using a cluster equipped with 8 NVIDIA H200 GPUs.

\paragraph{Evaluation Settings} We conduct evaluations on the curated Dentistry-Specific Benchmark (Section~\ref{sec:benchmark}), which assesses models across dental disease classification, lesion recognition, and common dental consultation scenarios. All models are required to provide responses to the given tasks. For models marked with an asterisk ($^*$), complex reasoning mode is enabled, while other models are instructed to directly output the correct option without additional reasoning steps.


\subsection{Results and Analysis}

As shown in Table~\ref{tab:main-result}, DentalGPT delivers clear and consistent performance gains over both comparable and substantially larger MLLMs across all expert-annotated datasets. It exhibits substantial improvements over its backbone, \texttt{Qwen2.5-VL-7B-Instruct}, underscoring the effectiveness of the proposed data construction pipeline and domain-aligned training strategy. Despite its compact 7B parameter scale, DentalGPT also surpasses many general-purpose models with over 100B parameters across nearly all benchmarks, demonstrating that domain-specialized modeling can achieve expert-level capability at a fraction of the computational cost and offering a promising path toward efficient, field-specific MLLMs.

Beyond expert-annotated datasets, DentalGPT maintains strong performance on the high-quality \textit{MMOral-OPG-Bench} and the \textit{DentalBench-Mixed} subset derived from general medical VQA benchmarks, outperforming most competing models and showing robust generalization across diverse dental tasks. Together, these results establish DentalGPT as a leading multimodal foundation model for dental image understanding.


\section{In-depth Analysis}

After evaluating the overall diagnostic performance, we further conducted fine-grained analyses to examine how each training stage enhances different capabilities of \textit{DentalGPT}, providing a clearer understanding of the respective roles of alignment and reinforcement learning in shaping its final diagnostic behavior.

\subsection{Effect of Multimodal Understanding Enhancement (on Stage I) }

Enhancement of multimodal understanding enables the MLLM to map its understanding of dental images into the textual semantic space, allowing it to effectively leverage existing knowledge for accurate interpretation of dentistry-specific multimodal tasks. To further assess how the strength of domain-specific alignment influences both visual comprehension and complex multimodal reasoning, we conducted an ablation study by varying the amount of Stage-I alignment data used during training.

\paragraph{Experimental Setup} Specifically, we conducted three controlled experiments by incorporating 0\%, 30\%, and 100\% of the Stage-I dataset to assess its impact. The effectiveness of different alignment levels was evaluated by analyzing reward improvements during the subsequent RL stage for complex reasoning. To ensure fairness, duplicate images between the RL data and alignment data were strictly removed, and all complex reasoning samples were excluded from the Stage-I dataset. Consistent with the \textit{DentalGPT} training setup, \texttt{Qwen2.5-VL-7B} was used as the backbone. Each model underwent 30 RL training steps, and accuracy-based reward changes on the validation set were monitored to assess multimodal reasoning performance.

\begin{figure}[ht!]
    \centering
    \includegraphics[width=0.6\textwidth]{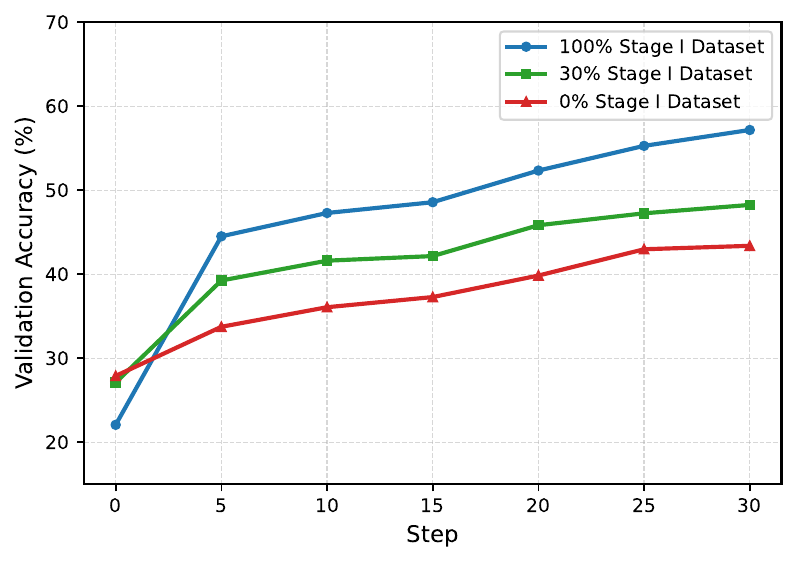}
    \caption{Accuracy reward (\%) of MLLM during RL training under different scales of the Stage I dataset.}
    \label{fig:indepth-midtraining-scaling}
\end{figure}

\paragraph{Results Analysis}
As shown in Figure~\ref{fig:indepth-midtraining-scaling}, the control group trained with 0\% Stage-I dataset, which receives no dental-domain knowledge, exhibits only marginal reward gains during the subsequent RL stage. In contrast, increasing the proportion of domain-specific training data consistently raises the performance ceiling throughout RL training. These results demonstrate that the Stage-I Training provides essential knowledge and visual grounding, thereby improving downstream reasoning capability in dentistry and enabling deeper image interpretation. In summary, incorporating the enhancement of multimodal understanding in dentistry effectively elevates the upper bound of performance attainable through multimodal complex reasoning.

\subsection{Effect of RL (on Stage II) }
This section investigates how the reinforcement learning (RL) stage further shapes the model's capabilities. After applying RL training on 10k multiple-choice dental questions, the performance of the model was reevaluated across the same set of benchmarks. As shown in Table~\ref{tab:indepth-posttraining-effect}, reinforcement learning brings consistent improvements across all tasks, demonstrating that it further enhances the model's ability to execute downstream dental tasks. These gains confirm that RL effectively strengthens both the accuracy and reliability of the model's reasoning in dental image understanding.

\begin{table}[ht!]
\centering
\small
\resizebox{0.7\textwidth}{!}{
\begin{tabular}{l|ccc}
\toprule
\textbf{Benchmarks} & \makecell{\textbf{Qwen2.5-VL}\\ \footnotesize{Backbone}} & \makecell{\textbf{Qwen2.5-VL}\\ \footnotesize{+ Stage I}  \\ \footnotesize{w/o Stage II}} & 
\makecell{\textbf{DentalGPT}\\ \footnotesize{w/ Stage I} \\ \footnotesize{\& Stage II}} \\
\midrule
MMOral-OPG-Bench & 27.0 & 56.8 & \textbf{60.0} \\
DentalBench-Mixed & 46.1 & 51.7 & \textbf{54.4} \\
Intraoral-Classification-I & 48.8 & 61.5 & \textbf{64.1} \\
Intraoral-Classification-II & 61.8 & 67.6 & \textbf{72.9} \\
Panorama-Classification & 50.0 & 78.4 & \textbf{84.0} \\
\midrule
Total & 46.7 & 63.2 & \textbf{67.1} \\
\bottomrule
\end{tabular}
}
\caption{Accuracy (\%) comparison between \texttt{Qwen2.5-VL}, \texttt{Qwen2.5-VL} with Stage I training, and DentalGPT (with both Stage I and Stage II training) across dentistry-specific benchmarks.}
\label{tab:indepth-posttraining-effect}
\end{table}

\subsection{Case Study}

\begin{figure}[ht!]
    \centering
    \includegraphics[width=\linewidth]{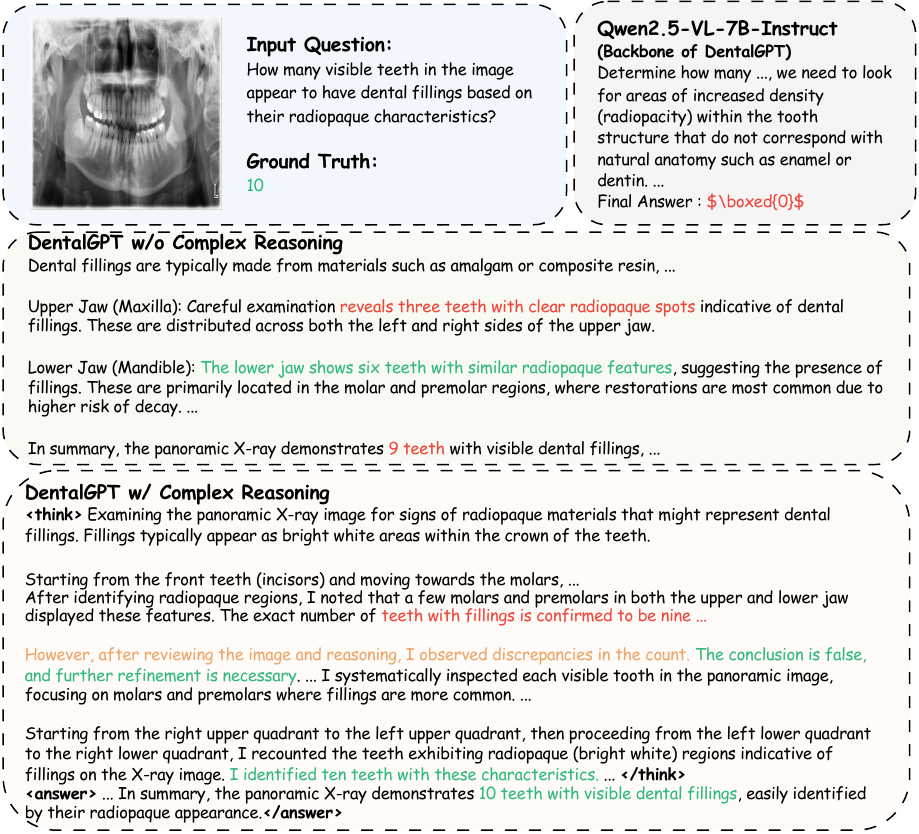}
    \caption{Examples of \textbf{DentalGPT} and its backbones analyzing a multimodal task in dentistry. Red indicates incorrect analysis, green indicates correct analysis, and orange highlights reflective turns in the complex reasoning process.}
    \label{fig:case-rl}
\end{figure}

We further analyze the outputs produced by \textbf{DentalGPT} at different training stages, as well as the original backbone model. As illustrated in Figure~\ref{fig:case-rl}, the backbone model struggles with this query: although it describes relevant visual characteristics in the image, it fails to correctly identify any teeth with fillings. After the enhancement of multimodal understanding, the model without complex reasoning is able to perform basic image analysis and detect most teeth, missing only one less noticeable case. This demonstrates the importance of alignment, which enables the model to learn how to interpret dental images and forms the foundation for further improving complex reasoning.

Finally, after all the training stages, \textbf{DentalGPT} with complex reasoning enabled conducts deeper multimodal analysis. Although intermediate thoughts may include incorrect counts, the iterative checking and reflection eventually lead to the correct answer. This again highlights the necessity of reinforcement learning for enhancing accurate multimodal reasoning in dentistry.

\section{Related Work}
Medical fields involve a large amount of imaging data, making them one of the most practical application areas for MLLMs. General medical MLLMs~\cite{huatuogptv,lin2025healthgptmedicallargevisionlanguage, lasateam2025lingshugeneralistfoundationmodel, Chao2024.01.18.24301503, yu2025medkgevalknowledgegraphbasedmultiturn} have demonstrated the ability to perform basic medical question answering and conduct preliminary visual analysis of medical images. Their aim is to generalize across diverse clinical scenarios and contribute to domain adaptation through large-scale multimodal medical data~\cite{zhang2023pmcvqa,liu2021slake,xie2024medtrinity,cai2025exploring}.

To better support complex imaging scenarios, subsequent research has focused on adapting MLLMs to specific medical modalities. Some models incorporate richer modality support beyond 2D images, extending to 3D medical scans~\cite{wu2023generalist, wu2025vision, wu2025towards} and biomedical signal analysis~\cite{zhao2024ecg,liu2024teach,chen2025shizhengpt,zeng2025wavemind}. Additionally, for tasks requiring high-resolution understanding such as reading detailed case images or zoom-level reasoning, several models~\cite{chen2025pathagent,wang2025pathology} introduce dedicated processing pipelines or tailored training strategies to address fine-grained clinical perception.

In dentistry, a specialty-centered medical domain, there has also been notable progress in multimodal diagnosis. \texttt{DentVLM}~\cite{meng2025dentvlm} aggregates large-scale hospital report data to build powerful image–text understanding models tailored for dental applications. \texttt{OralGPT}~\cite{hao2025oralgpt,oralgptomni} further advances this direction by evaluating multimodal benchmarks in dentistry and supporting various dental imaging modalities. These works demonstrate the growing recognition of dentistry as a valuable and distinct multimodal AI research scenario.


\section{Conclusion}
This work introduces \textbf{DentalGPT}, a specialized MLLM designed to address the challenges of multimodal diagnosis in dentistry. By constructing the largest annotated dental image dataset to date and integrating high-quality domain knowledge through a staged enhancement of multimodal understanding and reinforcement learning pipeline, the model gains the ability to capture fine-grained visual cues and perform more reliable disease-related reasoning. Extensive evaluations on intraoral, panoramic, and dental VQA benchmarks demonstrate that DentalGPT achieves strong performance despite its compact 7B parameter size, surpassing many state-of-the-art general-purpose MLLMs. These findings highlight the critical role of domain-specific data and training strategies in advancing dental AI, and underscore the potential of DentalGPT as a foundational model for future research and applications in automated dental imaging and intelligent oral healthcare.

\section*{Acknowledgment}
This work was supported by Shenzhen Medical Research Fund (B2503005), Major Frontier Exploration Program (Grant No. C10120250085) from the Shenzhen Medical Academy of Research and Translation (SMART), the Shenzhen Science and Technology Program (JCYJ20220818103001002), NSFC grant 72495131, Shenzhen Doctoral Startup Funding (RCBS20221008093330065), Tianyuan Fund for Mathematics of National Natural Science Foundation of China (NSFC) (12326608), Shenzhen Science and Technology Program (Shenzhen Key Laboratory Grant No. ZDSYS20230626091302006), the 1+1+1 CUHK-CUHK(SZ)-GDSTC Joint Collaboration Fund, Guangdong Provincial Key Laboratory of Mathematical Foundations for Artificial Intelligence (2023B1212010001), and Shenzhen Stability Science Program 2023.


\bibliography{ourbib}
\bibliographystyle{unsrt}


\end{document}